\documentclass[runningheads,american]{llncs}[2022/01/12]

\usepackage{amsmath,amssymb,amsfonts}
\usepackage{upquote}

\usepackage[american]{babel}

\usepackage[hyphens]{url}

\makeatletter
\g@addto@macro{\UrlBreaks}{\UrlOrds}
\makeatother

\usepackage[%
    rm={oldstyle=false,proportional=true},%
    sf={oldstyle=false,proportional=true},%
    tt={oldstyle=false,proportional=true,variable=false},%
    qt=false%
]{cfr-lm}

\usepackage[T1]{fontenc}

\usepackage[
  babel=true, %
  expansion=alltext,
  protrusion=alltext-nott, %
  final %
]{microtype}

\DisableLigatures{encoding = T1, family = tt* }

\usepackage{graphicx}

\usepackage{diagbox}

\usepackage{xcolor}

\usepackage{listings}

\definecolor{eclipseStrings}{RGB}{42,0.0,255}
\definecolor{eclipseKeywords}{RGB}{127,0,85}
\colorlet{numb}{magenta!60!black}

\lstdefinelanguage{json}{
    basicstyle=\normalfont\ttfamily,
    commentstyle=\color{eclipseStrings}, %
    stringstyle=\color{eclipseKeywords}, %
    numbers=left,
    numberstyle=\scriptsize,
    stepnumber=1,
    numbersep=8pt,
    showstringspaces=false,
    breaklines=true,
    frame=lines,
    string=[s]{"}{"},
    comment=[l]{:\ "},
    morecomment=[l]{:"},
    literate=
        *{0}{{{\color{numb}0}}}{1}
         {1}{{{\color{numb}1}}}{1}
         {2}{{{\color{numb}2}}}{1}
         {3}{{{\color{numb}3}}}{1}
         {4}{{{\color{numb}4}}}{1}
         {5}{{{\color{numb}5}}}{1}
         {6}{{{\color{numb}6}}}{1}
         {7}{{{\color{numb}7}}}{1}
         {8}{{{\color{numb}8}}}{1}
         {9}{{{\color{numb}9}}}{1}
}

\usepackage[autostyle=true]{csquotes}

\defineshorthand{"`}{\openautoquote}
\defineshorthand{"'}{\closeautoquote}

\usepackage{booktabs}

\usepackage{paralist}

\usepackage[%
  square,        %
  comma,         %
  numbers,       %
  sort&compress %
]{natbib}

\usepackage{etoolbox}
\makeatletter
\patchcmd{\NAT@test}{\else \NAT@nm}{\else \NAT@hyper@{\NAT@nm}}{}{}
\makeatother

\SetExpansion
[ context = sloppy,
  stretch = 30,
  shrink = 60,
  step = 5 ]
{ encoding = {OT1,T1,TS1} }
{ }

\usepackage[rflt]{floatflt}

\usepackage{pdfcomment}

\usepackage{stfloats}
\fnbelowfloat

\usepackage[group-minimum-digits=4,per-mode=fraction]{siunitx}
\addto\extrasgerman{\sisetup{locale = DE}}

\usepackage{hyperref}

\hypersetup{
  hidelinks,
  colorlinks=true,
  allcolors=black,
  pdfstartview=Fit,
  breaklinks=true
}

\usepackage[all]{hypcap}

\usepackage[caption=false,font=footnotesize]{subfig}

\usepackage{mindflow}

\usepackage[capitalise,nameinlink]{cleveref}

\crefname{section}{Sec.}{Sec.}
\Crefname{section}{Section}{Sections}
\crefname{listing}{List.}{List.}
\crefname{listing}{Listing}{Listings}
\Crefname{listing}{Listing}{Listings}
\crefname{lstlisting}{Listing}{Listings}
\Crefname{lstlisting}{Listing}{Listings}

\usepackage{lipsum}

\usepackage[math]{blindtext}
\usepackage{mwe}
\usepackage[realmainfile]{currfile}
\usepackage{tcolorbox}
\tcbuselibrary{listings}

\DeclareFontFamily{U}{MnSymbolC}{}
\DeclareSymbolFont{MnSyC}{U}{MnSymbolC}{m}{n}
\DeclareFontShape{U}{MnSymbolC}{m}{n}{
  <-6>    MnSymbolC5
  <6-7>   MnSymbolC6
  <7-8>   MnSymbolC7
  <8-9>   MnSymbolC8
  <9-10>  MnSymbolC9
  <10-12> MnSymbolC10
  <12->   MnSymbolC12%
}{}
\DeclareMathSymbol{\powerset}{\mathord}{MnSyC}{180}

\usepackage{xspace}

\makeatletter
\newcommand{\hydash}{\penalty\@M-\hskip\z@skip}
\makeatother

\input glyphtounicode
\usepackage{pifont}
\usepackage{graphicx}
\usepackage{stmaryrd}
\usepackage{pifont}
\usepackage{tablefootnote}
\usepackage{todonotes}
\usepackage{floatrow}
\usepackage{mathrsfs}
\usepackage{amsbsy} %
\usepackage{algorithm}
\usepackage{silence}
\usepackage{packages/algpseudocodex}
\usepackage{enumitem}
\usepackage{duckuments}
\usepackage[mathscr]{euscript}
\newfloatcommand{capbtabbox}{table}[][\FBwidth]

\long\def\IfNoTokA #1\IfNoTokB \IfNoTokB {}
\long\def\IfNoTokB \IfNoTokC #1#2{#1}    
\long\def\IfNoTokC           #1#2{#2}%

\long\def\IfNoTokens #1{\IfNoTokA\IfNoTokB #1\IfNoTokB\IfNoTokB\IfNoTokC }

\newcommand{\T}[1]{\boldsymbol{\mathscr{#1}}}   %

\newcommand{\added}[1]{#1}

\def\x{{\mathbf x}}

\def\R{{\mathbb{R}}}

\def\A{{\mathbf{A}}}
\def\B{{\mathbf{B}}}
\def\C{{\mathbf{C}}}

\def\M{{\mathbf{M}}}
\def\O{{\mathcal{O}}}

\newcommand{\V}[1]{\textbf{#1}}
\newcommand{\rev}[1]{#1}

\newcommand{\methodname}{\textsc{FRAPPE}}
\newcommand{\methodu}{\textsc{FRAPPE}}

\newcommand{\wtodo}[1]{{\textcolor{blue}{\textbf{[\IfNoTokens{#1}{TODO}{TODO: #1}]}}}}
\newcommand{\fnorm}[1]{\ensuremath{\left\lVert #1 \right\rVert_F}}

\newcommand{\zeros}{\textsc{Zeros}}

\newcommand{\col}[2]{#1^{(#2)}}
\newcommand{\rand}{\textsc{Rand}}

\newcommand{\cmark}{\ding{51}}%
\newcommand{\xmark}{\ding{55}}

\newcommand{\aboutascii}{\ensuremath{{\sim}}}

\begin{document}

\title{\texorpdfstring{FRAPPE: \underline{F}ast \underline{Ra}nk \underline{App}roximation with \underline{E}xplainable Features for Tensors}{FRAPPE: Fast Rank Approximation with Explainable Features for Tensors}}

\author{William Shiao\orcidID{0000-0001-5813-2266} \and
Evangelos E. Papalexakis\orcidID{0000-0002-3411-8483}}
\institute{University of California Riverside \\
\email{wshia002@ucr.edu, epapalex@cs.ucr.edu}}

\maketitle

\begin{abstract} \small\baselineskip=9pt
Tensor decompositions have proven to be effective in analyzing the structure of multidimensional data. However, most of these methods require a key parameter: the number of desired components. In the case of the CANDECOMP/PARAFAC decomposition (CPD), the ideal value for the number of components is known as the canonical rank and greatly affects the quality of the decomposition results. Existing methods use heuristics or Bayesian methods to estimate this value by repeatedly calculating the CPD, making them extremely computationally expensive. In this work, we propose \methodname{}, the first method to estimate the canonical rank of a tensor without having to compute the CPD. This method is the result of two key ideas. First, it is much cheaper to generate synthetic data with known rank compared to computing the CPD. Second, we can greatly improve the generalization ability and speed of our model by generating synthetic data that matches a given input tensor in terms of size and sparsity. We can then train a specialized single-use regression model on a synthetic set of tensors engineered to match a given input tensor and use that to estimate the canonical rank of the tensor---all without computing the expensive CPD. \methodname{} is over 24\texttimes{} faster than the best-performing baseline, and exhibits a $10\%$ improvement in MAPE on a synthetic dataset. It also performs as well as or better than the baselines on real-world datasets.

\begin{keywords}
Rank estimation \and Tensor decomposition \and Self-supervised \and CANDECOMP/PARAFAC
\end{keywords}
\end{abstract}

\section{Introduction}
\label{sec:intro}
Tensors are widely used in the field of computer science and can be used to model a variety of datasets. One popular family of methods that can be applied to tensors is tensor decompositions. \added{The performance of these methods typically depends on a key parameter---the number of components in the decomposition, also known as the rank of the decomposition. If our decomposition rank is too low, we may fail to capture key information, but if it is too high, we may capture too much noise. In the case of the popular CANDECOMP/PARAFAC decomposition, (CPD)~\cite{PARAFAC}, the ideal value for the rank of the decomposition is known as the canonical rank. However, finding this is a challenging task. A similar problem, finding the \textit{exact rank} of a tensor (the minimum number of CPD components required to represent the tensor exactly), is known to be NP-hard over the set of rational numbers~\cite{haastad1990tensor,hillar2013most}.}

As such, various heuristic-based approaches have been proposed to solve this task. One such method, proposed by \citet{bro2003new}, uses the core consistency diagnostic (CORCONDIA) to evaluate the ``appropriateness'' of a PARAFAC model. AutoTen~\cite{papalexakis2016autoten} uses CORCONDIA to select a suitable number of components automatically. Normalized SVD (NSVD)~\cite{Tsitsikas2020Decomposition} also attempts to provide a heuristic for the rank of a tensor, but it is not fully automatic and currently requires human supervision. \citet{normo} propose NORMO, which computes the CPD of a tensor at various ranks, looks at the mean correlation between each of its components, and selects the maximum number of components with no redundancy between the components.%

Another set of approaches is the Bayesian approaches to this problem. \citet{ZhaoZC14} attempt to tackle this problem by using Bayesian inference to automatically determine the decomposition's rank. \citet{morup2009automatic} use a Bayesian framework with Automatic Relevance Determination (ARD) to estimate the rank of a Tucker decomposition. This can easily be adapted since the CPD can be formulated as a restricted Tucker decomposition~\cite{tucker3}. \citet{rankCNN} compute the CPD and use a convolutional neural network (CNN) on the factors in an effort to determine the canonical rank of the tensor. However, they primarily focus on and evaluate their method within the image domain.

\added{%
One common theme across all of these methods is that they \textit{all require exhaustively computing the CPD multiple times across all the candidate ranks}. This is computationally expensive and impractical for large tensors, so we would like to avoid this step. However, doing so poses several challenges: (a) it is difficult to directly train a supervised machine learning model due to the scarcity of tensors with known canonical rank (which can typically only be determined via manual human inspection), and (b) it is difficult to develop a model capable of handling tensors from various domains and of various sizes.}

\added{To solve (a), we propose using synthetic tensors, which we can create with known canonical rank. To solve (b), we propose using hand-picked size-agnostic features extracted from each tensor. However, we find that supervised regression models trained using that data (see \cref{subsec:first-attempt}) fail to generalize well to real-world tensors. To overcome this limitation, we propose \methodname{}: a fully self-supervised model that uses synthetic data specifically tailored for a given input tensor. \methodname{} outperforms existing baselines on synthetic data and is on par with the best baselines on real-world data. It is also \aboutascii 24\texttimes{} faster than the best-performing baseline (NORMO).}

We evaluate the performance of both models on three classes of tensors: synthetic tensors, real tensors that have been extensively studied in existing literature to determine their rank, and the weight tensor of a convolutional neural network (to find the ideal CPD rank for compressing the weights while maintaining good classification accuracy). Finally, we also use these models to analyze which features are the best and worst predictors of a tensor's rank.

Our contributions include the following:

\begin{itemize}[noitemsep,topsep=1em]
	\item \textit{Novel method:} We propose \methodname{}---the first method to estimate the canonical rank of generic tensor \textit{without having to compute the CPD}.
	\item \textit{Speed:} We show that \methodname{} offers a \aboutascii{}24\texttimes{} improvement in evaluation time over the best-performing baseline method and is \aboutascii 1.6\texttimes{} faster than the fastest baseline.
	\item \textit{Extensive Evaluation:} We evaluate our methods and baseline methods on synthetically generated tensors, real tensors with known rank, and the weight tensor of a convolutional neural network. We show that \methodname{} outperforms the best baseline by about 10\% MAPE on a synthetic dataset and performs as well as or better than baselines on real-world datasets.
	\item \textit{Explainability:} We examine the importance of each of our hand-picked features to see which features contribute the most to classification results (see \cref{fig:feat_importances}).
	\item \textit{Reproducible:} We provide the Python code for our method online at \url{https://anonymous.4open.science/r/frappe-4D01/} (will be posted to publicly GitHub if accepted). We also provide the code to reproduce our baseline results.
\end{itemize}%
{%
\begin{table*}[!bt]
\caption{Comparison of different tensor rank estimation methods. All speeds are relative to the evaluation time of the slowest baseline method. Exact evaluation times are shown in \cref{table:main_table}. \dag: We did not benchmark against TRN-PD because it is designed for image data and image denoising, which is not the focus of this work.}
\label{table:salesman}
\centering
\resizebox{\linewidth}{!}{%
\begin{tabular}{r|c||c|c|c|c}
& \multicolumn{1}{c||}{\textbf{Proposed}} & \multicolumn{4}{c}{\textbf{Baseline Methods}} \\
\cline{2-6}
\textbf{} & \multicolumn{1}{l||}{\rotatebox[origin=c]{45}{\textbf{FRAPPE}}} & \multicolumn{1}{l|}{\rotatebox[origin=c]{45}{\textbf{NORMO}~\cite{normo}~ }} & \multicolumn{1}{l|}{\rotatebox[origin=c]{45}{\textbf{ARD-Tucker} \cite{morup2009automatic}}} & \multicolumn{1}{l|}{\rotatebox[origin=c]{45}{\textbf{AutoTen}~\cite{papalexakis2016autoten}}} & \multicolumn{1}{l}{\rotatebox[origin=c]{45}{\textbf{TRN-PD} \cite{rankCNN}}} \\ 
\hline
\textbf{Avoids Computing CPD} & \cmark & \xmark & \xmark & \xmark & \xmark \\
\textbf{Avoids Human Supervision} &  \cmark & \cmark & \cmark & \cmark & \cmark \\
\textbf{Avoids Strictly Labelled Dataset} & \cmark & \cmark & \cmark & \cmark & \xmark \\
\textbf{Explainable} & \cmark & $\circ$\tablefootnote{NORMO is partially explainable in that higher correlation in the factors indicates a bad choice for a given rank. However, NORMO does not provide any indication as to which attributes in the \textit{original} tensor contribute to higher/lower predictions.} & \xmark & \xmark & \xmark \\
\textbf{Works for General Tensors} & \cmark & \cmark & \cmark & \cmark & \xmark\tablefootnote{TRN-PD is designed for images and was originally evaluated for CPD-based denoising.} \\ 
\hline
\textbf{Eval. Speed} & $24\times$ & $1\times$ & $12\times$ & $18\times$ & \dag
\end{tabular}
}
\end{table*}}%
\section{Background}
\label{sec:background}
Here we provide an overview of the notation and background for this work. We denote a set, tensor, matrix, vector, and scalar as $\mathcal{X}$, $\T{X}, \textbf{X}$, $\textbf{x}$, $x$, respectively. We write the Frobenius norm of $\textbf{X}$ as $\left\lVert \textbf{X} \right\rVert_F$. Let $\T{T}_{(i, j, k)}$ denote the value in the $i$-th, $j$-th, and $k$-th entry along the 1st, 2nd, and 3rd modes, respectively, of the 3rd order tensor $\T{T}$. Let $(:)$ refer to all possible indexes in the given mode when used in tensor indexing (e.g., $\T{T} = \T{T}_{(:,:,:)}$ for a third-order tensor $\T{T}$).%

We also define several functions. Let $\zeros{}(\T{T})$ return the number of zero values in a tensor $\T{T}$, and $\rand(i,j)$ return a random matrix of size $i \times j$. Finally, let $\T{T} \sim \mathcal{N}_{i \times j \times k}(\mu, \sigma)$ represent sampling a tensor $\T{T}$ of size $i \times j \times k$ from a normal distribution with a mean of $\mu$ and standard deviation of $\sigma$. $\circ$ represents the tensor outer product.
\subsection{CANDECOMP/PARAFAC}
One of the most common tensor decompositions is the canonical polyadic or CANDECOMP/PARAFAC decomposition (CPD)~\cite{cpd1,cpd2}. The CPD decomposes a tensor $\T{T}$ into the sum of $R$ outer products of vectors. Formally, for a third-order tensor:%
\begin{equation}
    \T{T} \approx \sum^R_{r=1} \mathbf{a}_r \circ \mathbf{b}_r \circ \mathbf{c}_r
\end{equation}
\noindent We can then attempt to iteratively minimize the Frobenius norm of the error: $\left\lVert \T{T} - \sum^R_{r=1} \mathbf{a}_r \circ \mathbf{b}_r \circ \mathbf{c}_r \right\rVert_F$. A common method for this is to use the alternating least squares algorithm. By convention, we refer to the factors as matrices $\mathbf{A}, \mathbf{B}, \mathbf{C}$, which consist of the corresponding vectors horizontally stacked.  For example, the $i$-th column of $\mathbf{A}$ would be $\mathbf{a}_i$. A key component of this method is $R$---the \textit{rank of the decomposition}. As described above, choosing a good value for $R$ is extremely difficult. The ideal value for $R$ is the canonical rank of $\T{T}$, and estimating that value is the focus of this work.

\section{Problem Formulation}
\label{subsec:problem}
\added{We tackle the problem of predicting the canonical rank of a tensor. Formally, the problem can be stated as follows:}%
\begin{tcolorbox}[colback=white, colframe=black, title=Problem Definition]
\added{Given a tensor $\T{X} = \T{L} + \T{N}$, where $\T{L}$ is a tensor with low rank and $\T{N}$ is tensor of high-rank noise, find the canonical rank $R$ of $\T{X}$ such that we minimize $\left\lVert \T{L} - \sum^R_{r=1} \mathbf{a}_r \circ \mathbf{b}_r \circ \mathbf{c}_r \right\rVert_F$ where $\V{A},\V{B},\V{C}$ are factors from the CPD of $\T{X}$.}
\end{tcolorbox}
\noindent \added{Note that $R$ is different from the \textit{exact rank} of the tensor (the minimum number of components needed to exactly represent $\T{X}$), which is out of the scope of this work.}

\added{Another secondary goal of our method is to identify which features are important in predicting the rank of a tensor, which could help us develop/improve rank prediction methods in the future. We define this task as follows:} 

\begin{tcolorbox}[colback=white, colframe=black, title=Subproblem: Explainability]
\added{Given a tensor $\T{X}$ and a predicted rank $\hat{R}$, compute the importance scores $\textbf{t}$ of different attributes of $\T{X}$ in the regression result.}
\end{tcolorbox}

\section{Proposed Method}
\label{sec:method}
\label{subsec:proposed_method}
We first describe which hand-picked features we extract from the tensor and the reasoning behind selecting each one. We then train a regression model on the dataset and show that it has some key limitations: (a) it performs well on the synthetic dataset but fails to generalize well to real data, and (b) it requires a relatively expensive training step. We then propose our method, $\methodname{}$, which addresses both of these issues.

\subsection{Tensor Features}
\label{subsec:features}
Given a tensor $\T{T} \in \mathbb{R}^{I \times J \times K}$, we extract 112 features (4D or higher tensors require some adjustment, as described in \cref{subsec:higher_order}): \\

\begin{itemize}[topsep=0.2em]
    \item \textbf{Tensor Dimensions (\#1--3):} $I, J, K$. \rev{We include these features so that the underlying model can adjust the expectations for other parameters relative to the size of the tensor. This allows our model to generalize to different input sizes.}

    \item \textbf{Non-zeros (\#4--13):} The total number of non-zero values in the tensor, as well as the minimum, median, and maximum number of non-zero values over all possible slices of the tensor. That is, we extract the number of non-zeros from $\T{T}_{(i, :, :)} \forall i \in [1, I]$, $\T{T}_{(:, j, :)} \forall j \in [1, J]$, and $\T{T}_{(:, :, k)} \forall k \in [1, K]$. The intuition behind this feature is that, generally, the rank of a tensor increases as it becomes more sparse. Measuring this for all possible slices allows us to essentially measure the sparsity along each dimension.

    \item \textbf{Ranks of Slices (\#14-22):} The minimum, median, and maximum rank over all possible slices of the tensor. The intuition behind this feature is that if slices along any given two dimensions are generally of low rank (implying that fewer components are needed in the SVD), fewer components may also be required in the CPD.

    \item \textbf{Thresholded Non-zeros (\#23--103):} The minimum, median, and maximum number of values in the \textit{normalized} (we use min-max normalization) tensor less than or equal to each of the following thresholds: $[0.1,0.2,\ldots, 0.9]$ across all possible slices of the tensor. This is an adaptation of the number of non-zeros (features \#4-13) for dense tensors and allows us to find the number of low-value entries, which could be a result of noise and be of lesser importance to the canonical rank of the tensor.
    \item \textbf{Correlations (\#104--112):} The minimum, median, and maximum correlation over all possible pairs of slices of the tensor. The intuition behind this feature is that higher correlation between slices generally indicates that fewer components would be needed along a given axis, resulting in a lower canonical rank.
\end{itemize}

\vspace{\baselineskip}

\noindent Note that the number of features is constant, \textit{regardless of the size of the tensor}. This allows it to easily scale to large tensors. We then train a LightGBM~\cite{ke2017lightgbm} model on these features, with the goal of predicting the rank of the target tensor. %

\subsubsection{Why Correlation?}
\label{subsubsec:why_correlation}%
\rev{%
It may not be immediately apparent why we choose to use the correlation between different slices. We choose to do so because the distribution of pairwise correlation of vectors in a matrix is a good indicator of its rank. We show this theoretically (under some assumptions) below. We hypothesize that this remains true in the tensor case; while difficult to formally prove, we also show that this feature is effective in our model in \cref{fig:feat_importances}.}

\begin{proof} \label{proof:corr}
\rev{%
We first define the rank of a matrix $\M$ as $k$, where $k$ is the largest value such that the $k$-th singular value $\sigma_k > \epsilon$. Let us assume that any two vectors $\V{x},\V{y}$ has an outer product $\V{x} \V{y}^T$ that can be written as $\V{a} \V{b}^T + \V{N}$, where $\V{N} = \gamma \sigma_1 \V{u}_1\V{v}_1^T + \gamma^2 \sigma_2 \V{u}_2\V{v}_2^T + \ldots + \gamma^k \sigma_k \V{u}_k \V{v}_k^T$ such that $\sigma_i < \epsilon\ \forall i$, where $\epsilon$ is an arbitrarily small number and $\textsc{Corr}(\V{x}, \V{y}) \rightarrow 1 \implies \gamma\rightarrow 0$. Let $c = \mathbb{E}\left[ \textsc{Corr}(\M^{(i)}, \M^{(j)}) \right] \forall (\M^{(i)}, \M^{(j)}) \in \mathscr{C}$. Let $\mathbf{M} \in \R^{m \times n}$, and $\col{\M}{i}$ be the $i$-th column of $\M$. %
For the sake of simplicity, let us also assume that a set $\mathscr{C}$ contains all $\frac{n}{2}$ non-overlapping pairs of columns, and all other pairs have a correlation of 0. Then, we can write $\M$ as $\V{a}_1\V{b}_1^T + \V{a}_2\V{b}_2^T + \ldots + \V{a}_{n/2}\V{b}_{n/2}^T + \V{N}_1 + \V{N}_2 + \ldots + \V{N}_{n/2}$. Let $\V{N}_\Sigma = \V{N}_1 + \V{N}_2 + \ldots + \V{N}_{n/2}$. Since we have at most $\frac{n}{2}$ duplicate components, the maximum value for a singular value of $\V{N}$ is $\frac{n}{2} \gamma_{c} \epsilon$, which is still less than $\epsilon$ for $\gamma < \frac{2}{n}$, and subsequently true for sufficiently large $c$. Therefore, the rank of $\M$ decreases as our mean pairwise correlation $c$ increases.}%
\end{proof}

\subsubsection{Computational Complexity.}
Features \#1--13 and \#23--103 are extremely quick to calculate, requiring only a single pass over the tensor. Features \#14--22 and \#104--112 are significantly more expensive since they require computing the SVD and the correlation, respectively, over various slices of the tensor. However, these operations are still much faster than the baseline methods, which require repeatedly computing the full CPD over the whole tensor. \cref{table:main_table} compares the speed of our method against the baselines.

Let $n$ $=\max(I, J, K)$. Features \#1-13 take a linear amount of time to calculate with respect to the number of non-zero values in the tensor, with a worst-case of $\O(n^3)$ (in the case of a dense tensor). Features \#14-22 take $\O((I + J + K)s) = \O(ns)$ time, where $s$ is the runtime of sparse SVD on a slice of our tensor. In the case of a fully dense tensor, this takes $\O((I + J + K) n^3) = \O(n^4)$ time to calculate.

Features \#23--103 require the tensor to be min-max normalized first, which can be done in $\O(n^3)$ time. It also takes $\O(n^3)$ to count the number above/below each threshold. Features \#104--112 take $\O(n^2)$ time to calculate over $\frac{n (n-1)}{2}$ pairs, resulting in a runtime of $\O(n^4)$.

Overall, these features take a \textit{maximum} of $\O(n^4)$ time to calculate for dense tensors, where $n$ is the size of the largest dimension in the tensor. It is worth noting that it would take significantly less time for sparse tensors (e.g., a multi-view graph, knowledge base, word coccurrence). This is also a massive improvement compared to the runtime of computing the CPD once, let alone the repeated runs required by existing methods. We show a comparison of the runtimes below in \cref{table:main_table}.%
\subsubsection{Extension to Higher-Order Tensors}
\label{subsec:higher_order}
While the features described above work well for 3rd-order tensors, they require some modification for 4th or higher-order tensors. For example, the rank features are computed over all slices of the tensor. In the 3rd order case, these slices are matrices, but in the 4th (or higher) order case, these slices are 3rd (or higher) order sub-tensors. Finding the rank of a matrix can be done in polynomial time via the SVD, but finding the rank of a 3D tensor is the problem we are tackling in this paper. Due to this issue and the relatively low importance of the rank features in \cref{fig:feat_importances}, we omit the rank features for the 4D model and find that it still performs fairly well. \rev{This is shown empirically in our experiments on a 4D AlexNet weight tensor below in \cref{subsec:cnn_eval}.} As with existing work in canonical rank estimation~\cite{normo,papalexakis2016autoten}, we focus on primarily on 3rd-order tensors and reserve detailed evaluation for future work.
\subsection{A First Attempt}
\label{subsec:first-attempt}
\added{We can now train a regression model using the features defined in \cref{subsec:features} above. To do this, we first generate a synthetic dataset with tensors of known rank (see \cref{subsec:synth_dataset}). We then train a LightGBM~\cite{ke2017lightgbm} model on this synthetic dataset and evaluate the model on each of the evaluation datasets in \cref{sec:experiments} below. However, we find that the LightGBM model is unable to generalize well beyond the synthetic datasets and incorrectly predicts the rank on all of the real-world datasets (described in \cref{subsec:real_world}). This is likely because real-world tensors come from a variety of different domains.}

\added{Another limitation of this approach is that we have to train our model on a very large dataset to ensure that we account for different ranks, sparsities, and noise levels. While this makes predicting the rank of a new tensor very quick, it greatly increases the amount of time required to train a regression model. To address this, we propose \methodname{}.}
\renewcommand{\algorithmicrequire}{\textbf{Input:}}%
\renewcommand{\algorithmicensure}{\textbf{Output:}}%
\begin{algorithm}[!ht]
    \caption{Pseudocode for \methodu{}: a self-supervised algorithm to estimate the rank of a given tensor. Note that comments precede the line that they describe.}
    \label{algo:self_supervised}
    \begin{algorithmic}[1]
        \Require{An input tensor $\T{T} \in \mathbb{R}^{i \times j \times k}$, $\rho$ (the maximum rank to consider), $\tau$ (the number of times to sample from each rank), and $\alpha$ (the maximum amount of noise to add in synthetic samples).}
        \Ensure{$\hat{R}$, an estimate for the rank of the input tensor $\T{T}$.}
        \Function{\methodu{}}{$\T{T}$,\ $\rho,\ \tau,\ \alpha$}
            \State $\texttt{data} \leftarrow [\ ]$
            \State $\texttt{ranks} \leftarrow [\ ]$
            \LComment{Calculate the sparsity of our tensor}
            \State $s \leftarrow \frac{\Call{zeros}{\T{T}}}{i\times j \times k}$
            \For{$x \in [1,\ \tau]$}
                \For{$r \in [1, \rho]$}
                    \State $\A \leftarrow \Call{Sparsify}{\Call{Rand}{i, r}, s}$
                    \State $\B \leftarrow \Call{Sparsify}{\Call{Rand}{j, r}, s}$
                    \State $\C \leftarrow \Call{Sparsify}{\Call{Rand}{k, r}, s}$
                    \LComment{Construct a synthetic tensor $\T{S}$ from $\A,\B,\C$}
                    \State $\T{S} \leftarrow \sum^r_{y=1} \mathbf{a}_y \circ \mathbf{b}_y \circ \mathbf{c}_y$
                    \LComment{Select a random noise ratio.}
                    \State $n \sim [0, \alpha]$
                    \LComment{Sample an $i\times j\times k$ Gaussian noise tensor}
                    \State $\T{N} \sim \mathcal{N}_{i \times j \times k}(0, 1)$%
                    \LComment{Scale the noise and add it to $\T{S}$}
                    \State $\T{S} \leftarrow \T{S} + (n\frac{\fnorm{\T{S}}}{\fnorm{\T{N}}}) \T{N}$
                    \LComment{Extract features from $\T{S}$, as described in \cref{subsec:features}}
                    \State \hspace{-0.20em}$\texttt{data}.append(\Call{ExtractFeatures}{\T{S}})$ 
                    \State $\texttt{ranks}.append(r)$
                \EndFor
            \EndFor

            \State Train a LightGBM regression model on $\texttt{data}$ and $\texttt{ranks}$
            \State Evaluate the model on $\Call{ExtractFeatures}{\T{T}}$ to get our prediction $\hat{r}$
            \State \Return $\hat{R}$
        \EndFunction
    \end{algorithmic}
    \vspace{1em}
    \begin{algorithmic}[1]
        \Require{An input matrix $\M \in \mathbb{R}^{a \times b}$ and the desired sparsity level $s$.}
        \Ensure{The sparsified matrix.}
        \Function{Sparsify}{$\M, s$}
            \State Select $s \times a \times b$ random indices of $\M$
            \State Set each of these indices to 0
            \State \Return $\M$
        \EndFunction
    \end{algorithmic}
\end{algorithm}
\subsection{\methodu{}}
\label{subsec:self-supervised}

\added{The primary issue we need to solve with \methodname{} compared to the approach described in \cref{subsec:first-attempt} is the poor generalization of the naive regression model. This poor generalization is fundamentally caused by a distribution difference between the training data and the input tensor. We can address this issue by ensuring that the synthetic training data closely mimics a given input tensor. We do this by attempting to match the size and sparsity of the input tensor when generating the synthetic dataset. We can then train a specialized LightGBM model \textit{for each tensor being considered}. This specialized model only performs evaluation on the input tensor once before being discarded.}

\added{Generating training data engineered to mimic an input tensor also solves the issue of long training times since less data is required to achieve the same results. This brings down training times significantly, allowing us to train a disposable regression model for each input tensor while maintaining a speed advantage over the other baseline models. Note that the ``training'' happens at \textit{evaluation} time, and no external training step is required. \methodu{} is formally described in detail in \cref{algo:self_supervised}.}

{\setlength{\belowcaptionskip}{-18pt}
\begin{figure}[ht]
    \centering
    \includegraphics[width=\linewidth]{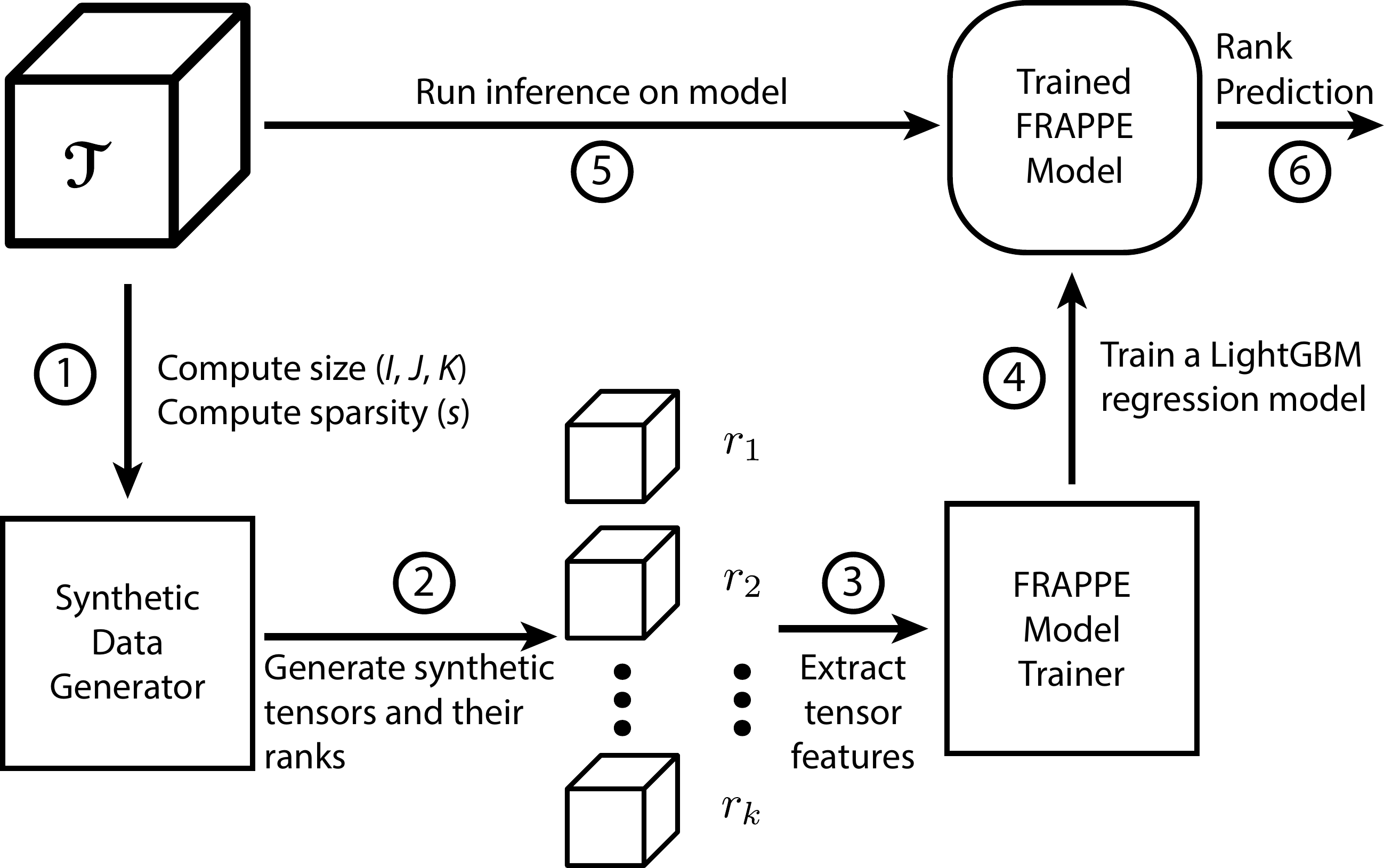}
    \caption{Diagram of \methodu{}. We first generate synthetic tensors based on the size and sparsity of the input tensor, extract features, train the model, and use that model to predict the rank of the original input tensor.}
    \label{fig:frappe_diagram}
\end{figure}
}

\section{Experimental Evaluation}
\label{sec:experiments}

Evaluating the effectiveness of our methods is difficult because there are very few tensors with known rank so we cannot solely rely on real-world data. As such, we choose to evaluate our methods on three different types of data:

\begin{enumerate}
    \item \textbf{Synthetic Tensors:} Predicting the rank of tensors that we generate to have known rank (described in \cref{subsec:synth_dataset}).
    \item \textbf{Real-world Tensors:} Predicting the rank of real-world tensors that have been previously studied and have known rank (described in \cref{subsec:real_world}).
    \item \textbf{CNN Weights:} Compressing the \rev{4D} weights of an AlexNet~\cite{alexnet} model's final layer (described in \cref{subsec:cnn_eval}).
\end{enumerate}

We compare the performance and runtime of our methods to other current state-of-the-art methods. All of the timing experiments are conducted on a \texttt{c3-highcpu-176} Google Cloud Platform instance with 176 CPU cores and 352 GB of RAM.

\subsection{Baseline Methods}
\label{subsec:baselines}

We compare \methodname{} and \methodu{} against three baselines: NORMO~\cite{normo}, AutoTen~\cite{papalexakis2016autoten} and ARD-Tucker~\cite{morup2009automatic}. These methods require a maximum rank to consider. Since this greatly affects the runtime of the algorithm, we used double the true rank of the tensor for the two algorithms. This theoretically bounds the maximum error, but setting it higher would greatly increase both the runtime and computational cost of running the baselines. We also use this same bound for \methodu{}.

We had to slightly modify AutoTen to only use the Frobenius-norm-based evaluation because we have negative values in our dataset, which causes the KL-divergence-based evaluation to fail. This slightly reduces the accuracy of AutoTen but greatly improves its runtime. We also had to modify NORMO and AutoTen to add support for 4D tensors in order to run it on our CNN weight tensor. We chose to use the suggested value of $\delta=0.7$ for NORMO.%

\begin{table}[!htp]
\caption{Performance of the rank estimation methods on real-world datasets. The first column is the known rank of the datasets, and correct estimates are bolded. We can see that \methodu{} and AutoTen generally perform the best overall.}
\label{table:real_world}
\centering
\resizebox{\linewidth}{!}{%
\begin{tabular}{l|cccccc} 
\toprule
\multicolumn{1}{l|}{\textbf{Dataset}} & \multicolumn{1}{c}{\textbf{Rank}} & \multicolumn{1}{c}{\textbf{FRAPPE}} & \multicolumn{1}{c}{\textbf{NORMO}} & \multicolumn{1}{c}{\textbf{AutoTen}} & \multicolumn{1}{c}{\textbf{ARD}} \\
\midrule
\texttt{amino\_acid} \cite{broAmino} & 3 \cite{broAmino,kiersAmino} & \textbf{3} & 4 & \textbf{3} & \textbf{3} \\
\texttt{dorrit} \cite{jackknife} & 4 \cite{jackknife} & \textbf{4} & \textbf{4} & \textbf{4} & \textbf{4} \\
\texttt{sugar} \cite{broSugar} & 3 \cite{broSugar} & \textbf{3} & 4 & 2 & \textbf{3} \\
\texttt{reality\_mining} \cite{realityMining} \hspace{10pt} & 6?\tablefootnote{The number of communities in this dataset was explored in \cite{realityMiningClustering} and estimated to be 6. The number of communities loosely corresponds to the canonical rank of the tensor, but this is not a guarantee like with the fluorescence datasets.} \cite{realityMiningClustering} & 9 & 2 & 2 & 18 \\
\texttt{enron} \cite{enronDataset} & 4/7? \cite{enronDedicom,conceptDriftTensor} & 3 & 1 & \textbf{4} & 12 \\
\bottomrule
\end{tabular}
}
\end{table}
\vspace{-0.25in}
\subsection{Synthetic Dataset}
\label{subsec:synth_dataset}

To generate a synthetic dataset, we first generate the CPD factor matrices $\mathbf{A}, \mathbf{B}, \mathbf{C}$ by drawing values from a uniform random distribution. We ensure that we generate a diverse dataset by generating a set percentage of tensors from each of the following subsets (each of which corresponds to a common real-world tensor):

\vspace{0.5em}
\begin{enumerate}
    \item \textbf{Sparse tensors with sparse noise}: Tensors that have sparsity introduced in the factors and Gaussian noise added to non-zero elements in the tensor. \textit{Example:} a weighted time-evolving graph.
    \item \textbf{Sparse tensors with dense noise}: Tensors that have sparsity introduced in the factors and Gaussian noise added to all elements in the tensor.\\\textit{Example:} a time-evolving grid of sensor readings.
    \item \textbf{Dense tensors}: Tensors that have fully dense tensors and dense noise added. \textit{Example:} a tensor of fluorometer readings.
\end{enumerate}
\vspace{0.5em}

Each of these cases represents a different type of tensor that can be found in real-world datasets. We add 2\%--10\% noise in each case and scale the amount of noise by the norm of the noise tensor compared to the generated tensor. More formally, where $\alpha$ is the noise ratio, $\T{N}$ is our noise tensor and $\T{G}$ is our generated tensor: $\T{G} = \T{G} + \alpha \frac{\fnorm{\T{G}}}{\fnorm{\T{N}}} \T{N}$. This is similar to the noise addition process in the Tensor Toolbox's~\cite{tensortoolbox} \texttt{create\_problem} function.
\begin{figure}[!htp]
    \centering
    \resizebox{\linewidth}{!}{
    \input{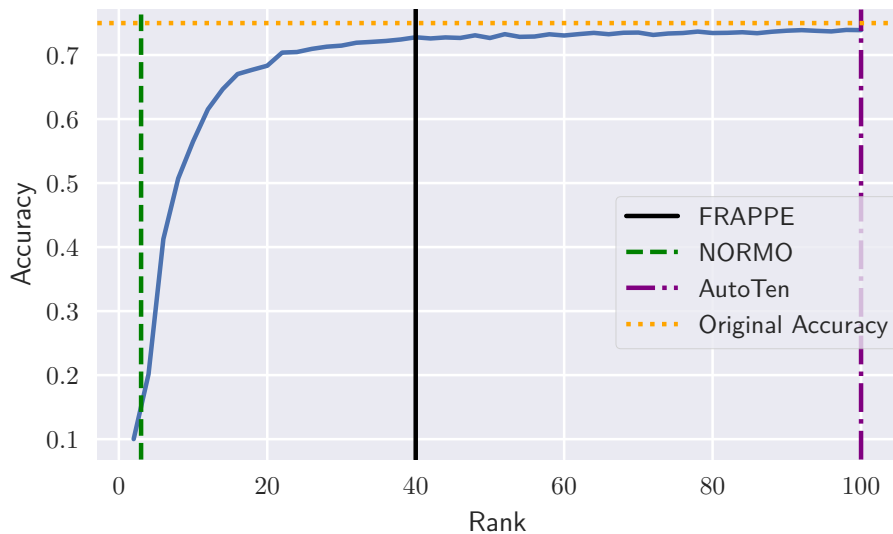}
    }
    \vspace{-0.25in}
    \caption{Plot of AlexNet performance with the final CNN layer compressed at various ranks. Also shown are the rank estimates produced by \methodname{} and the baselines. NORMO predicts a very low value and AutoTen predicts a very high value. \methodname{} and \methodu{} pick what appear to be reasonable values. ARD-Tucker did not terminate in time and is not included.}
    \label{fig:alexnet_perf}
\end{figure}%

\subsection{Real-World Datasets}
\label{subsec:real_world}

We evaluate the effectiveness of our methods and baseline methods on datasets of known rank. These include three fluorescence datasets: \texttt{amino\_acid}~\cite{broAmino}, \texttt{dorrit}~\cite{jackknife}, and \texttt{sugar}~\cite{broSugar}. These datasets contain naturally tri-linear data with a known number of chemicals, each of which corresponds to a component in the CPD. We also evaluate our methods on the \texttt{reality\_mining}~\cite{realityMining} and \texttt{enron}~\cite{enronDataset} datasets. The canonical rank of these tensors is less clear, as we have to rely on existing literature that has manually analyzed the data and determined the approximate rank.%

From the results displayed in \cref{table:real_world}, we can see \methodu{} is able to perfectly estimate the rank of the fluorescence datasets. ARD is also able to correctly estimate the rank of the datasets. NORMO and AutoTen perform fairly well but are slightly off. Only AutoTen is able to successfully estimate the rank for the \texttt{enron} dataset, but FRAPPE is also very close. Unfortunately, we are limited in the amount of data we can use for this evaluation because there are very few real-world tensors with known ranks. While it is difficult to draw any concrete conclusions due to the limited number of datasets, we can see that \methodu{} performs just as well as or slightly better than other existing rank estimation methods.%
\subsection{CNN-Based Evaluation}
\label{subsec:cnn_eval}
We evaluate the performance of our network in selecting the correct number of components when decomposing a 4D weight tensor for a convolutional neural network (CNN). We focus specifically on the last layer ($\R^{256 \times 256 \times 3 \times 3}$) of an AlexNet \cite{alexnet} model trained on the CIFAR-10 \cite{cifar10} image dataset.

The goal of this task is to select the correct number of components such that we compress the weights as much as possible and maintain as high accuracy as possible. This rank would be the ``knee'' of the rank-accuracy curve. A plot of this curve with the values selected by the various rank estimation methods is shown in \cref{fig:alexnet_perf}.

NORMO predicts a low estimate of 3 for the rank of the weight tensor, and AutoTen predicts the maximum estimate of 100. Both of these appear to be poor choices for this task since the accuracy is very low at rank 3 and the gains are heavily diminished by rank 100. \methodu{} estimates the rank to be 40, which is not perfect, but appears to be fairly reasonable estimates near the ``knee'' of the plot. ARD-Tucker did not terminate within 24 hours and is therefore excluded from the plot.

\begin{figure}[!htp]
\begin{floatrow}
\ffigbox{%
  \includegraphics[width=\linewidth]{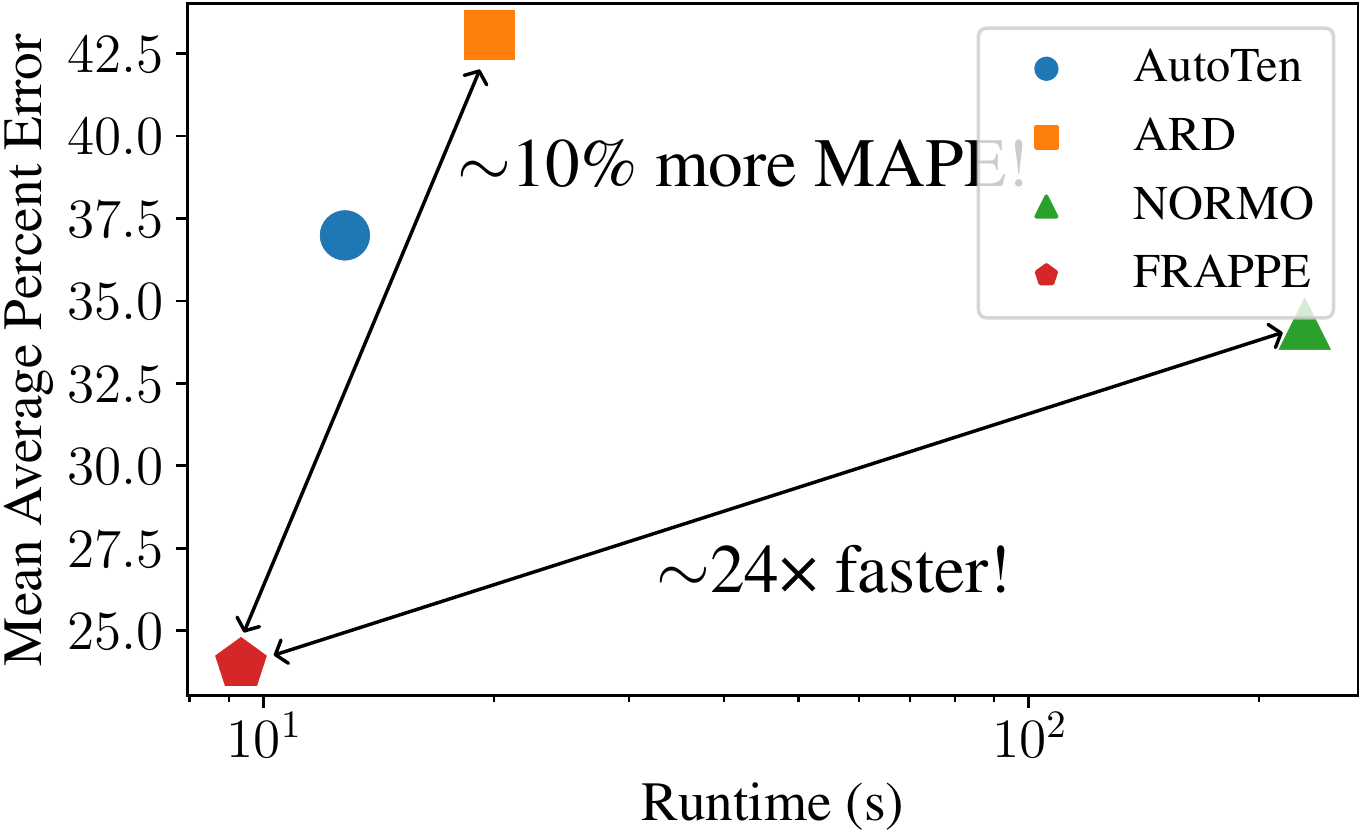}%
}{%
  \caption{Runtime/performance comparison between the different rank estimation methods on a synthetic dataset. The closer to the lower left, the better.}%
}
\capbtabbox{%
\resizebox{\linewidth}{!}{%
    \begin{tabular}{lcccc}
    \toprule
    \textbf{Model} & \textbf{MAPE} & \textbf{MAE} & \textbf{MSE} & \textbf{Runtime} \\ 
    \midrule
    AutoTen\tablefootnote{Note that we only used the Frobenius-norm-based AutoTen for these synthetic examples due to the presence of negative values in the tensors (causing KL-divergence-based AutoTen to fail). This could result in lower accuracy but also decrease the evaluation times.} \cite{papalexakis2016autoten} & $36.99\%$ & 11.55 & 314.80 & \ \ 12.77s \\
    ARD \cite{morup2009automatic} & $43.06\%$ & 13.52 & 414.70 & \ 19.72s \\
    NORMO \cite{normo} & $34.36\%$ & 10.44 & 282.38 & 229.34s \\
    \methodu{} & $\textbf{23.99\%}$ & \ \textbf{6.43} & \ \textbf{81.40} & \ \ \ \textbf{9.34s} \\
    \bottomrule
    \end{tabular}%
}
}{%
  \caption{Mean Absolute Percentage Error (MAPE), Mean Absolute Error (MAE), and Mean Square Error (MSE), and mean runtime of different models on a synthetic dataset.}%
  \label{table:main_table}%
}%
\end{floatrow}%
\end{figure}%
\subsection{Synthetic Dataset Results}
\label{subsec:synth_results}

\cref{table:main_table} shows that \methodname{} outperforms the baseline methods, with a \aboutascii 10\% improvement in Mean Absolute Percentage Error (MAPE) over NORMO (the best-performing baseline), \aboutascii 4 rank improvement in Mean Absolute Error (MAE), and an improvement of 200 in Mean Squared Error (MSE). It is also \aboutascii 24$\times$ faster than NORMO, and \aboutascii 1.6\texttimes{} faster than AutoTen (the fastest baseline).

\subsection{Feature Importance Results}%
\begin{figure}[!htp]
    \centering
    \resizebox{\linewidth}{!}{
    \input{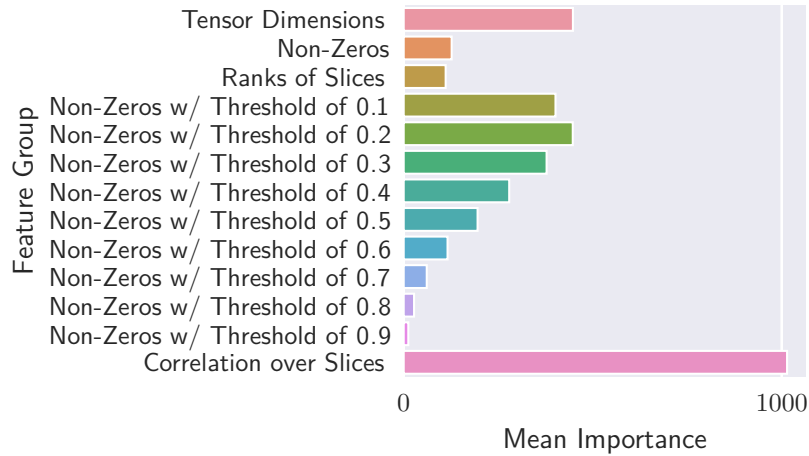}
    }%
    \vspace{-0.35in}
    \caption{Chart of feature group importance (higher is more important), as output by a LightGBM model trained on a synthetic dataset. The most important feature group is the correlation over slices of the tensor, followed by the number of non-zeros thresholded at 0.2. We calculate feature group importance by how often a feature is used in splits. The feature groups are described in more detail in \cref{subsec:features}.}
    \label{fig:feat_importances}
\end{figure}

\label{subsec:feat_results}%
In \cref{fig:feat_importances}, we plot the importance of the different feature groups (described in \cref{subsec:features} above). We compute the importance of a feature based on the number of times it is used in the split of a LightGBM tree.

We can see that the most important features are the correlation over different slices. This makes intuitive sense because higher correlation between slices along a given mode means that components are more related along that mode, leading to fewer components being required. This also happens to be the intuition behind NORMO, one of the baseline methods, which looks instead at the correlation between different factors of the CPD.

Another interesting set of features are the thresholded non-zero counts. In \cref{fig:feat_importances}, we can see that the importance increases from a threshold of 0.1 to 0.2, but steadily decreases after that until 0.9, which is almost irrelevant. This is likely because lower relative values (since we normalize the values) are more likely to be noise artifacts or simply unimportant values in the decomposition. Lower values are also more likely to be excluded in an $R$-rank decomposition, simply because the incremental error from excluding that value is smaller than it would be for a larger-valued element.

\section{Additional Related Work}
\label{sec:related}

On top of the CPD rank estimation methods described above, there have also been similar methods developed for different tensor factorizations. \citet{UnsupervisedTensorReactive} propose NTF$k$, which is an unsupervised method to perform Non-negative Tensor Factorization (NTF). NTF$k$ uses a custom clustering algorithm (similar to $k$-means) to cluster the columns of a factor matrix and evaluates their quality with silhouette score. It then uses a combination of those scores and the reconstruction error to choose a good target for the number of components.
\citet{Nebgen2020AFactorization} attempt to solve a similar problem for non-negative matrix factorization (NMF), which decomposes a matrix $\mathbf{V}$ into $\mathbf{V} = \mathbf{W} \mathbf{H}$.

They first repeatedly calculate the NMF on data perturbed by noise from a random uniform distribution. Then, they use $k$-means to cluster the columns of $\mathbf{W}$. They then use the Akaike information criterion (AIC) and silhouette score to calculate the quality of the clusters. Finally, they use a neural network over a sliding window of the AIC and cluster silhouette values to determine the ideal number of components.

\section{Conclusion}
\label{sec:conclusions}
\added{In this work, we propose \methodname{}, a novel method for estimating the canonical rank of a tensor in a self-supervised manner without having to compute the expensive CPD. The key insight of this approach is that it is much cheaper to generate synthetic data of known rank than to compute the actual CPD. Following this idea, we also propose a set of hand-crafted features that can effectively determine a given tensor's rank.}

\added{We find that naively training a supervised regression model on this data results in poor generalization to real-world tensors and requires a long training step. To alleviate this issue, we propose \methodname{}, which first generates synthetic tensors (with known rank) from an input tensor and trains a new model on these synthetic tensors, improving the generalization ability of the model, reducing the amount of training data required, and speeding up the model training.}

\added{We extensively evaluate \methodname{} on three different types of data: synthetic tensors, real-world tensors, and the weight tensor of a CNN. We find that our model generally outperforms existing baselines across all three datasets, is over 24\texttimes{} faster than the best-performing baseline (NORMO), and \aboutascii{}1.6\texttimes{} faster than the fastest baseline (AutoTen).}

\added{Finally, we examine the features we used to make these predictions and why we believe these features are important in estimating a tensor's rank. We provide both theoretical reasoning and empirical evidence for why the pair-wise correlation between slices is a good indicator of the tensor's rank. We believe this opens up an interesting line of research about estimating the number of components of a tensor decomposition without ever having to compute a single decomposition.}

\section{Ethical Statement}
\label{sec:ethics}

One important ethical consideration regarding this work can arise if it is used on sensitive information like financial or healthcare data. As with all machine learning methods, our method is not infallible and may produce incorrect results. In certain mission-critical systems, that can potentially lead to severe consequences if care is not taken to validate model results.

Another ethical concern is the possibility of adversarial attacks on the model. In our work, we do not consider the case of adversarially crafted input tensors designed to trick the model into performing poorly or predicting an extremely high rank, which could potentially result in maliciously altered results.

\renewcommand{\bibsection}{\section*{References}} %
\bibliographystyle{splncs04nat}
\begingroup
  \microtypecontext{expansion=sloppy}
  \small %
  \bibliography{bib/refs,bib/vagelis_refs,bib/mendeley}
\endgroup

\end{document}